# Investigating Knowledge Distillation Through Neural Networks for Protein Binding Affinity Prediction

Wajid Arshad Abbasi[1, *], Syed Ali Abbas[1], Maryum Bibi[1], Saiqa Andleeb[2], Muhammad Naveed Akhtar[3]

[1]Computaional Biology and Data Analysis Lab., Institute of Computing, King Abdullah Campus, University of Azad Jammu & Kashmir, Muzaffarabad, AJ&K, 13100, Pakistan.

[2]Biotechnology Lab., Department of Zoology, King Abdullah Campus, University of Azad Jammu & Kashmir, Muzaffarabad, AJ&K, 13100, Pakistan.

[3]Department of Computer and Information Sciences(DCIS), Pakistan Institute of Engineering and Applied Sciences(PIEAS), Islamabad, Pakistan.

**\*Corresponding author**: Wajid A. Abbasi (e-mail: wajidarshad@uajk.edu.pk).

## Abstract

The trade-off between predictive accuracy and data availability makes it difficult to predict protein–protein binding affinity accurately. The lack of experimentally resolved protein structures limits the performance of structure-based machine learning models, which generally outperform sequence-based methods. In order to overcome this constraint, we suggest a regression framework based on knowledge distillation that uses protein structural data during training and only needs sequence data during inference. The suggested method uses binding affinity labels and intermediate feature representations to jointly supervise the training of a sequence-based student network under the guidance of a structure-informed teacher network. Leave-One-Complex-Out (LOCO) cross-validation was used to assess the framework on a non-redundant protein–protein binding affinity benchmark dataset. A maximum Pearson correlation coefficient ($P_r$) of 0.375 and an RMSE of 2.712 kcal/mol were obtained by sequence-only baseline models, whereas a $P_r$ of 0.512 and an RMSE of 2.445 kcal/mol were obtained by structure-based models. With a $P_r$ of 0.481 and an RMSE of 2.488 kcal/mol, the distillation-based student model greatly enhanced sequence-only performance. Improved agreement and decreased bias were further confirmed by thorough error analyses. With the potential to close the performance gap between sequence-based and structure-based models as larger datasets become available, these findings show that knowledge distillation is an efficient method for transferring structural knowledge to sequence-based predictors. The source code for running inference with the proposed distillation-based binding affinity predictor can be accessed at https://github.com/wajidarshad/ProteinAffinityKD.

**Keywords:** Protein–protein binding affinity, Knowledge distillation, Sequence-based prediction, Structure-informed learning, Deep learning regression

# Introduction

Protein binding affinity represents the intensity of interaction between a protein and its ligand, usually a small molecule or another protein(Kastritis and Bonvin, 2013). It is a key measure in biochemistry and pharmacology, as it dictates how tightly the ligand binds to a protein and for what duration the complex will remain intact(Marsh and Teichmann, 2015). Its measurement is in the form of a dissociation constant ($K_d$), which is the concentration of the ligand at which a protein will have half of its binding sites occupied, or the Gibbs free energy($\Delta G$), which can be calculated from the $K_d$ measurement(Moal and Fernández-Recio, 2012). Quantitative assessments of the binding affinity of proteins are crucial for several reasons(Kastritis and Bonvin, 2013). First, it could aid in comprehending the operational mechanisms of the protein, in addition to the processes through which it binds with other proteins in a cell. Second, if drugs are required to bind with high affinity to the target proteins, then the affinity could aid in the development of novel drugs(Kairys et al., 2019). Finally, the affinity of binding of proteins could furnish details regarding the specificity of the protein and ligand interaction, which is needed for the development of selective activators and inhibitors(Huggins et al., 2012).

There are several experimental methods to determine protein binding affinity. Each of them has advantages as well as disadvantages. One popular method is Surface Plasmon Resonance or SPR. This method is based on measuring, in real-time, protein-ligand interaction by detecting changes in the surface plasmons induced by the interaction on the surface of a sensor chip(Tang et al., 2010). This method is highly sensitive. This method is capable of measuring information regarding the interaction, including its kinetic and thermodynamic components. This method is costlier and needs special equipment. Another popular method is Isothermal Titration Calorimetry or ITC. This method is based on measuring heat evolving or absorbing during protein-ligand interaction(Saponaro, 2018). It will be able to correctly quantify the affinity as well as the thermodynamic values, but the technique requires a large amount of protein. It might also take a large amount of time. Techniques such as fluorescence resonance energy transfer, FRET, or fluorescence polarization, FP, could be used to estimate the affinity values for the proteins(Moerke, 2009; Sekar and Periasamy, 2003). These methods might be faster and more economical compared to the SPR or the ITC methods. These might sometimes require tagging the proteins with fluorescent compounds. The principal drawbacks of the experimental methods make them incongruous in terms of broader applications in view of the cost, the availability of the samples, generation of artifacts, or the dynamic range.

To counter the above-mentioned limitations in the measurement of an experimental result by the experimental method, there have been the development of computational models that predict the binding affinity by molecular dynamics simulation and potential energy functions through machine learning algorithms(Audie and Scarlata, 2007; Guo and Yamaguchi, 2022a; Horton and Lewis, 1992; Panday and Alexov, 2022; Siebenmorgen and Zacharias, 2019; Su et al., 2009). MD simulations can be computationally expensive, especially for large proteins or complex systems and generate large amounts of data that can be complex and difficult to analyze.

Empirical energy functions may depend on the specific parameterization of the energy function and unable o account for protein flexibility. Among computational methods, machine learning-based predictive models can be the preferred choice to predict protein binding affinity as they treat all those complex factors involved in protein-protein interactions, that cannot be handled through manually-curated functions, implicitly (Ain et al., 2015).

In recent years, numerous machine learning–based approaches have been proposed for protein-protein binding affinity prediction, demonstrating significant improvements over traditional physics-based methods (Guo and Yamaguchi, 2022b). Existing methods can broadly be categorized into structure-based and sequence-based approaches. Structure-based methods exploit three-dimensional (3D) protein conformations to model spatial and physicochemical interactions and generally achieve superior predictive performance (Gainza et al., 2020; Liu et al., 2021; Rodrigues et al., 2019; Romero-Molina et al., 2022; Wee and Xia, 2022). However, their applicability is limited by the availability of experimentally resolved high quality protein structures, which are often unavailable for proteins encountered in real-world prediction scenarios(Charih et al., 2025). This dependency on structural information restricts the scalability and practical deployment of structure-based binding affinity predictors.

Sequence-based methods, in contrast, require only amino acid sequences and can therefore be applied universally (Abbasi et al., 2020; Yugandhar and Gromiha, 2014). Nevertheless, their prediction accuracy is generally poor compared to structure-based methods because the affinity of protein binding is primarily dictated by the 3D structural arrangement, not the protein sequence. The challenge in this regard is the use of the protein sequence information in modeling the complex spatial relationship. To compensate for this shortcoming, a knowledge distillation method, conceptualized based on the work of Hinton et al. (Hinton et al., 2015), is proposed. The method uses the protein 3D structural feature as a teacher model, with the protein sequence feature as the student model (see Fig. 1). It implicitly transplants the structural information into the student by the teacher during the process of training. In this way, the student still has the practical advantages of sequence-only inputs while improving predictive accuracy for a model without requiring protein structural information at inference time.

## Methods

In this section, we have presented the methodology employed to develop and evaluate the proposed knowledge distillation–based framework for protein–protein binding affinity prediction, in which structural information is leveraged during training through a teacher model to enhance the performance of a sequence-based student predictor.

### Dataset and Preprocessing

In this study, we used the Protein Binding Affinity Benchmark Dataset v2.0 (Kastritis et al., 2011), a subset of Docking Benchmark v4.0 (Hwang et al., 2010), comprising 144 non-redundant protein complexes with experimentally determined binding affinities and resolved bound and unbound 3D structures. Protein complexes in this dataset have known binding affinities in terms of binding free energy and disassociation constant. Following established preprocessing protocols, 128 heterodimeric complexes were retained after quality filtering, enabling the use of previously published structural and experimental descriptors(Abbasi et al., 2018; Moal et al., 2011; Yugandhar and Gromiha, 2014). This allows us to use descriptors from Moal et al. and Dias et al., (Dias and Kolaczkowski, 2017; Moal et al., 2011).

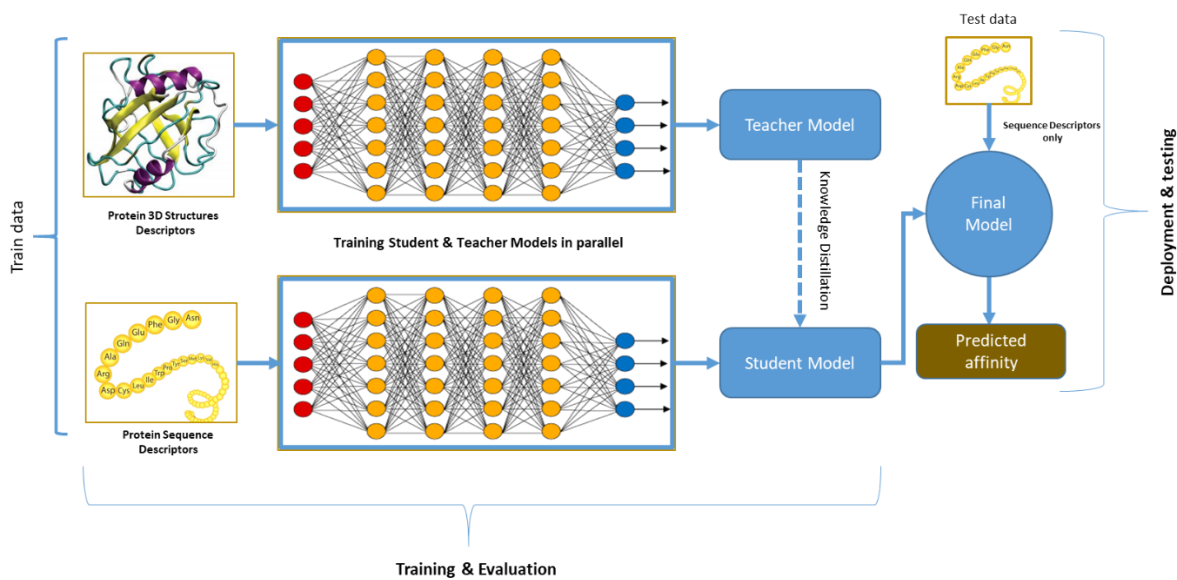

**Fig. 1. Overview of the knowledge distillation framework for protein–protein binding affinity prediction.** During training, a structure-based teacher network is trained using protein structural descriptors in parallel with training a sequence-based student network using protein sequence descriptors. A student model is supervised by both the ground-truth binding affinity labels and teacher's outputs and intermediate representations, thus allowing transferring structural knowledge. During the inference, only protein sequence descriptors are needed, while binding affinity prediction for a protein complex without resolved three-dimensional structures is allowed.

An external validation dataset comprising 39 protein–protein complexes with known binding free energies was used for stringent performance evaluation of the baseline and distillation models. The dataset was derived from Chen et al. by excluding complexes with more than two chains or chain lengths below 50 residues (Chen et al., 2013), and has been previously employed for independent validation in related studies (Abbasi et al., 2018; Moal and Fernández-Recio, 2014).

## Regression Models

### Learning Framework

We suggest a machine learning framework that uses a knowledge distillation paradigm to jointly leverage protein sequence and structural information in order to predict protein–protein binding affinities. The main concept of the suggested method is to take advantage of rich three-dimensional (3D) structural descriptors during training while keeping a sequence-only requirement at inference time (See Fig.1). This allows proteins without resolved structures to be practically deployed. The basic trade-off between predictive accuracy and data accessibility that is frequently seen in binding affinity predictors based on structure and sequence is addressed by this design.

The proposed framework is made up of two neural networks: a student model and a teacher model. Protein structure-based descriptors, which provide high predictive accuracy by encoding specific geometric, energetic, and physicochemical interaction information, are used to train the teacher model. Concurrently, the student model is trained using protein sequence-based descriptors, which are less expressive but widely accessible. In training, the teacher imparts knowledge to the student by minimizing both its own supervised regression loss and a distillation loss that motivates the student to imitate the teacher's predictions and/or internal representations. The student model can implicitly learn structural features from sequence information alone thanks to this knowledge distillation technique, which is based on the formulation presented by Hinton et al., (Hinton et al., 2015). Consequently, only protein sequence descriptors are needed for testing, and the trained student model achieves better binding affinity prediction performance.

The training objective and knowledge distillation loss used to optimize the teacher and student models are described in detail in the upcoming subsection.

*Model Architecture*

Here, we begin by presenting the binding affinity prediction problem as a regression problem. In the proposed framework of machine learning based binding affinity prediction, our dataset $D = \{(x_i^s, x_i^t, y_i)\}_{i=1}^{N}$ consisting of N examples, where $x_i^s$ represents protein complex level sequence-based descriptors, $x_i^t$ represents protein complex level structure-based descriptors, and $y_i$ is the experimentally measured binding affinity. Let $f_t(\cdot)$ denotes the teacher network and $f_s(\cdot)$ denotes the student network. The proposed framework follows a teacher–student learning paradigm in which two neural networks are trained using different types of protein descriptors but optimized under a unified distillation objective. The teacher network operates on structure-based descriptors of protein–protein complexes, while the student network uses sequence-based descriptors. Both networks are designed for continuous binding affinity prediction, formulated as a regression task.

The Teacher Network (Structure-Based Model) is designed to leverage rich structural information and therefore employs a higher-capacity architecture. It consists of a fully connected feedforward neural network composed of multiple linear layers followed by nonlinear activations. Given a structure-based descriptor vector $x^t \in R^{d_t}$, the teacher network learns a hierarchical latent representation:

$$h^t = \psi_t(x^t)$$

where $\psi_t(\cdot)$ denotes the feature extraction component of the teacher. The final regression output is produced by a linear prediction head:

$$y'^t_i = f_t(x^t_i)$$

The higher representational capacity of the teacher enables it to capture complex spatial, energetic, and physicochemical interactions encoded in three-dimensional protein structures.

The Student Network (Sequence-Based Model) operates exclusively on sequence-derived descriptors and is therefore intentionally designed as a lighter-weight model. It follows a similar fully connected architecture but with fewer layers and parameters to reflect the reduced information content of sequence-only features. Given a sequence-based descriptor vector $x^s \in R^{d_s}$, the student network computes:

$$h^s = \Phi_s(x^s)$$

where $\Phi(\cdot)$ denotes the feature extraction component of the student. The final regression output is produced by a linear prediction head:

$$y'^s_i = f_s(x^s_i)$$

All models were implemented in PyTorch(Paszke et al., 2019), and architectural choices were adapted to the dimensionality and nature of the corresponding descriptors while keeping the distillation interface consistent across models.

*Loss Function and Distillation Interface*

As baseline both the $f_t(\cdot)$ and $f_s(\cdot)$ are trained using mean squared error (MSE) as supervised regression loss with respect to the ground truth binding affinity:

$$\mathcal{L}^t_{sup} = \frac{1}{N}\sum_{i=1}^{N}(y'^t_i - y_i)^2$$

$$\mathcal{L}^s_{sup} = \frac{1}{N}\sum_{i=1}^{N}(y'^s_i - y_i)^2$$

To transfer knowledge from the teacher to the student, a distillation loss is introduced that encourages the student to mimic the teacher's predictions:

$$\mathcal{L}_{dist} = \frac{1}{N}\sum_{i=1}^{N}(y'^s_i - y'^t_i)^2$$

The teacher predictions are treated as a fixed target for the student during backpropagation.

To transfer richer structural knowledge beyond predictions, we incorporate feature-level distillation, where intermediate representations of the teacher guide the student. The student is encouraged to align its latent representation with that of the teacher using the following mean squared error loss:

$$\mathcal{L}_{feat} = \frac{1}{N} \sum_{i=1}^{N} (h_i^s - h_i^t)^2$$

During optimization, gradients from this loss are propagated only to the student network, while the teacher features are treated as fixed targets.

Following two different loss functions have been used to optimize the student network in the distilation setup.

$$\mathcal{L}_{total} = (1 - \lambda_{out})\mathcal{L}_{sup}^s - \lambda_{out}\mathcal{L}_{dist}$$

and

$$\mathcal{L}_{total} = (1 - \lambda_{out})\mathcal{L}_{sup}^s - \lambda_{out}\mathcal{L}_{dist} - \lambda_{feat}\mathcal{L}_{feat}$$

Where $\lambda_{out}$ and $\lambda_{feat}$ control the contribution of each distillation component. During optimization, gradients from this loss are propagated only to the student network, while the teacher features are treated as fixed targets. The teacher network is optimized independently using $\mathcal{L}_{sup}^t$.

### Training and Inference Strategy

During training, both teacher and student networks are optimized simultaneously. The teacher network is trained using supervised regression loss based on experimentally measured binding affinities, while the student network is optimized using a combination of supervised loss and distillation losses derived from the teacher's outputs and intermediate representations. Importantly, the teacher network is not frozen, allowing co-adaptation during training.

At inference time, only the student network is used, requiring sequence information alone, thereby enabling binding affinity prediction for protein complexes without resolved three-dimensional structures.

### Implementation Details

All models were implemented using the PyTorch framework. The architectural design of the teacher and student networks, including the number of layers, hidden dimensions, activation functions, and the selection of the distillation layer, was adapted to the dimensionality and characteristics of the corresponding structure-based or sequence-based descriptors. A generic architectural overview is provided in supplementary Table S1. Complete architectural specifications, training hyperparameters, and optimization settings are reported in the supplementary Tables S2 and S3 to ensure full reproducibility of the experiments.

## Protein Descriptors

In this work, we employ both sequence-based and 3D structure-based descriptors within a knowledge distillation framework. Sequence-based descriptorsconstitute the student input space and are available during both training and inference, whereas structure-based descriptors are treated as privileged information and are accessible only during training through the teacher model. All feature representations are standardized to zero mean and unit variance across the dataset. Details of the individual feature representations are provided below.

*Protein 3D Structure Descriptors*

Proteins perform their functions through interactions that depend on their three-dimensional (3D) structures. As a result, a protein complex's structural properties play a critical role in determining its binding affinity. We employed several complex-level feature representations to extract relevant protein complex structural properties. We used these features as both baseline inputs and for knowledge distillation. The following outlines the different kinds of structural feature representations used with a consistent (identical) length throughout all protein complexes in our dataset.

**Dias descriptors**: Dias and Kolaczkowski's study on protein–protein binding affinity prediction provided the protein complex descriptors used in this investigation(Dias and Kolaczkowski, 2017). The pH of the binding assay, the experimental temperature, and the experimental technique used to measure the binding affinity for a particular protein complex are all included in these features, which summarize the main experimental parameters under which the binding affinities were evaluated(Dias and Kolaczkowski, 2017). Because this type of experimental metadata is known to affect the reported affinities, it offers crucial extra information that makes it easier to interpret experiments in the context of predictive models. The experimental method's categorical string features were transformed into feature vectors using binary one-hot encoding, as is customary, to make numerical computation easier(Harris and Harris, 2012). For every protein complex in the dataset, these experimental and assay-related characteristics come together to create a 26-dimensional feature representation. The utility of the feature set has been confirmed by earlier research, which found a Pearson correlation coefficient of 0.68 between the measured and predicted binding affinities using the provided feature set(Dias and Kolaczkowski, 2017).

**Moal descriptors**: These descriptors, which offer a 200-dimensional representation of protein complexes and capture interface characteristics and conformational changes upon binding, were taken from the study by Moal et al., (Moal et al., 2011). Statistical potentials, solvation and entropy terms, unbound–bound energy differences, and other interaction energies like electrostatics, dispersion, exchange repulsion, and hydrogen bonding are all included in the feature set. For the benchmark dataset, a Pearson correlation coefficient of 0.55 between predicted and experimental binding affinities has been reported using these descriptors (Moal et al., 2011).

**Number of interacting residue pairs (NIRP):** Non-covalent interactions between amino acid residues at the binding interface of the ligand and receptor proteins are primarily responsible for stabilizing interactions within a protein–protein complex(Zhu et al., 2008). The complex's binding mode and binding free energy are significantly influenced by the particular makeup and spatial arrangement of these interacting residues (Swapna et al., 2012). Inspired by this finding, we calculated the frequencies of non-redundant residue–residue pairs throughout the protein–protein interface, treating residue pairs A:B and B:A as equivalent, in order to characterize interfacial interactions. These frequencies were derived from the bound 3D structures of the ligand (L) and receptor (R) proteins using a distance cutoff of 8 Å to define interfacial contacts. This approach yields a 211-dimensional feature representation that captures the statistical distribution of interfacial residue interactions within a protein complex. This feature representation has previously been shown to be effective for protein–protein binding affinity prediction and was successfully employed in Abbasi *et al.*(Abbasi et al., 2018).

***Blosum (Interface):*** As discussed earlier, BLOSUM-based features were extracted to capture the substitution patterns of physicochemically similar amino acids within protein sequences, reflecting evolutionary conservation relevant to protein function. BLOSUM matrices encode empirically derived substitution scores obtained from aligned protein families and are widely used to represent sequence similarity in a biologically meaningful manner. In addition to whole-sequence representations, we computed BLOSUM features specifically for interface residues of protein complexes. Interface residues were defined as amino acids having at least one atom within a distance cutoff of 8 Å from any atom of the interacting partner protein in the complex. This interface-focused representation emphasizes residues directly involved in binding interactions and enables the model to incorporate localized sequence information that is most relevant to protein–protein binding affinity prediction.

***Protein Sequence Descriptors***

To model the sequence-based characteristics of a protein–protein complex comprising ligand and receptor chains, sequence-derived features were first computed independently for each protein chain. For complexes containing multiple chains on either the ligand or receptor side, chain-level feature vectors were aggregated by averaging across all corresponding chains, resulting in a single representative feature vector for the ligand and a single vector for the receptor, respectively. This aggregation strategy enables a fixed-length representation while preserving the overall sequence composition of multi-chain partners. Subsequently, the ligand and receptor feature vectors were concatenated to form a unified sequence-based feature representation of the protein complex, consistent with previously established approaches (Abbasi et al., 2020; Ahmad and Mizuguchi, 2011). Detailed descriptions of the individual chain-level sequence-based feature descriptors employed in this study are provided in the following subsections.

***k-mer Composition (k-mer):*** The frequency of consecutive amino acid subsequences of length $k$ is used by a well-known sequence descriptor called k-mer composition to characterize a protein sequence(Leslie et al., 2002). This representation captures local sequence patterns and compositional biases that reveal information about protein function and interaction propensity. In this study, we employed dipeptide (2-mer) composition, counting the instances of each of the 20 standard amino acids' potential ordered pairs along the protein sequence. The resulting counts were normalized by sequence length to account for differences in protein size. This approach generates a 400-dimensional feature vector for each protein chain that provides a condensed but informative summary of sequence composition relevant to protein–protein binding affinity prediction.

***Group k-mer Composition (k-mer-G):*** We extracted grouped k-mer composition features from protein sequences in order to capture sequence composition while incorporating biochemical similarity among amino acids. Using this method, the twenty naturally occurring amino acids were first divided into seven groups according to their electrostatic and hydrophobic characteristics, as detailed in (Shen et al., 2007). In addition to accounting for physicochemical similarity between residues, this grouping lowers feature dimensionality. The frequencies of contiguous k-mers made from these amino acid groups were then calculated, with k ranging from 2 to 4, in order to encode protein sequences. The resulting feature representation is compact and captures local sequence patterns at several scales. This encoding generates a feature vector of length $7^k$ for a given value of k, providing fixed-length representations appropriate for binding affinity prediction based on machine learning.

***BLOSUM-62 Descriptors(Blosum):*** For capturing both the amino acid compositional features as well as the substitution pattern of the protein sequences, we used BLOSUM-based features. For this purpose, each protein sequence is represented as a vector of dimension 20 by taking the mean of the column vectors of a BLOSUM matrix defined based on the presence of amino acids in the sequence. For this purpose, each column of the BLOSUM matrix holds empirically measured substitution scores based on the probability of an amino acid being replaced by other amino acids of similar physicochemical properties in evolutionary processes. For this purpose, in this work, the popular BLOSUM-62 substitution matrix was used to calculate these features (Eddy, 2004). This representation has been successfully applied in several previous studies and has demonstrated effectiveness for modeling protein sequence similarity and interaction-related properties (Abbasi et al., 2020, 2018; Aumentado-Armstrong et al., 2015; Westen et al., 2013; Zaki et al., 2009).

***ProPy Descriptors (ProPy):*** To effectively represent biophysical characteristics of amino acids as well as structural information derived from protein sequences, we used the feature extraction tool ProPy(Cao et al., 2013). The ProPy tool produces a comprehensive feature set, ProPy descriptors, which is 1537 dimensional. This feature set encompasses several feature classes, namely pseudo-amino acid composition (PseAAC), which is a type of feature set involving sequence order information, and amino acid composition, transition, and distribution features. Additionally, it encompasses other feature classes, namely, sequence-order coupling numbers, quasi-sequient order features, and features based on structural and physicochemical attributes. These feature classes represent comprehensive structural information about protein sequences and have been frequently utilized as features for protein sequence prediction(Abbasi et al., 2020; Li et al., 2006; Limongelli et al., 2015).

***Position-Specific Scoring Matrix Descriptors(PSSM):*** To capture the evolutionary relationships as well as the conserved patterns in the protein sequences, we used Position-Specific Scoring Matrix (PSSM) based descriptors. A PSSM is a representation of the evolutionary profile of a protein that calculates the probability of the presence of each of the 20 amino acids in the protein sequence. The PSSM was generated for each protein chain in the complex using PSI-BLAST with three iterations for the non-redundant (nr) protein database with the E-value threshold of $10^{-3}$(Altschul et al., 1997; Pruitt et al., 2005). To derive a constant-length feature, regardless of sequence variability, a PSSM was condensed using a calculation of averages of column vectors over a sequence, which generated a 20-element feature vector for a protein chain(Abbasi et al., 2020).

***ProtParam features (ProtParam):*** In order to incorporate the important physicochemical features from the protein sequence into our work, we have resorted to the usage of the "ProtParam tool" from the ExPASy server to extract data on molecular weight, aromaticity, instability index, isoelectric point, and fractions of secondary structure(Abbasi et al., 2020; Gasteiger et al., 2005; Guruprasad et al., 1990). The usage of the tool gives us a 7-dimensional vector representation for the input protein sequence.

## Model validation, selection and performance assessment

We employed Leave-One-Complex-Out (LOCO) cross-validation to evaluate the performance of the proposed regression models on the non-redundant binding affinity benchmark dataset (Kastritis et al., 2011). Under this scheme, the model is trained on N−1 protein complexes and tested on the remaining complex, and this procedure is repeated until each complex has been used once as the test sample (Abbasi and Minhas, 2016). Model performance was assessed using standard regression metrics, including Pearson correlation coefficient ($P_r$), root mean squared error (RMSE), and mean absolute error (MAE). To check the statistical significance of the results, we have also estimated the P-value of the correlation coefficient scores. Final performance scores were obtained by averaging the metrics over three independent LOCO runs with different data shufflings, and the averaged results are reported in the Results and Discussion section.

## Results and Comaprison

In this section, we present a comprehensive evaluation of the proposed knowledge distillation–based regression framework for protein–protein binding affinity prediction. The performance of the sequence-based student model, trained with and without distillation, is systematically compared against the structure-based teacher model and conventional baseline regressors. All models are evaluated using Leave-One-Complex-Out (LOCO) cross-validation on the non-redundant binding affinity benchmark dataset, ensuring a rigorous and unbiased assessment. Prediction accuracy is assessed using correlation-based and error-based regression metrics, allowing quantitative analysis of the contribution of structural knowledge transfer to sequence-only binding affinity prediction. The results highlight the effectiveness of the proposed distillation strategy in improving predictive performance while retaining the practical advantage of requiring

**Table 1.** Baseline performance of the student regression network trained independently, without teacher guidance or knowledge distillation, on protein sequence descriptors using leave-one-cluster-out cross-validation over the affinity benchmark dataset.

| Sequence Descriptors | MAE | RMSE | $P_r$ | P-value |
|---|---|---|---|---|
| k-mer | 2.312 | 2.819 | 0.321 | 0.001 |
| k-mer-G | 2.138 | 2.751 | 0.358 | 0.001 |
| Blosum | 2.080 | 2.664 | 0.333 | 0.002 |
| **Propy** | **2.090** | **2.712** | **0.375** | **0.000** |
| PSSM | 2.189 | 2.791 | 0.254 | 0.004 |
| ProtParam | 2.181 | 2.742 | 0.244 | 0.006 |

Bold faced values indicate best performance for each model.
MSE: Mean Absolute Error, RMSE: Root Mean Sequred Error, $P_r$: Pearson correlation coefficient

only protein sequence information at inference time. In the following sections, we present a comprehensive evaluation of the proposed framework and compare its performance against sequence-based baselines to demonstrate the effectiveness of the distillation strategy.

**Baseline Binding Affinity Prediction Using Protein Sequence Descriptors**

We first evaluated baseline regression models trained exclusively on sequence-derived descriptors to establish a lower-bound reference for protein–protein binding affinity prediction. These models rely solely on sequence information and therefore represent the most broadly applicable prediction setting.

Using the LOCO cross-validation protocol, the sequence-based baseline models achieved a maximum Pearson correlation coefficient ($P_r$) of 0.375, an RMSE of 2.712 kcal/mol, and an MAE

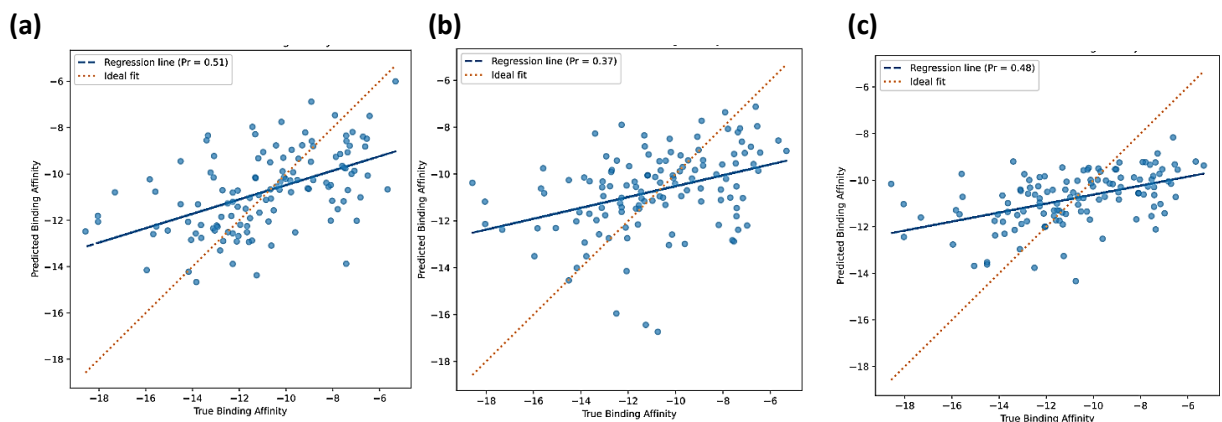

**Fig. 2.** Scatter plots showing the comparison of the experimental and predicted protein protein binding affinities using propy and Moal descriptors on LOCO CV. **(a)** structure-based baseline model, **(b)** sequence-based baseline model, and **(c)** knowledge distillation–based model. Each plot shows the linear regression fit (dashed line) and the ideal identity line (dotted line). The sequence-based model exhibits moderate correlation, while the structure-based model achieves higher predictive accuracy. The distillation-based student model substantially improves over the sequence-only baseline and approaches the performance of the structure-based model, demonstrating effective transfer of structural knowledge during training.

**Table 2.** Upper-bound performance of the teacher regression model trained on protein structure descriptors, representing the ideal supervision signal in the knowledge distillation framework and evaluated using LOCO cross-validation over the affinity benchmark dataset.

| Structure Descriptors | MAE | RMSE | $P_r$ | P-value |
|---|---|---|---|---|
| NIRP | 2.000 | 2.480 | 0.481 | 0.001 |
| **Moal descriptors** | **1.845** | **2.445** | **0.512** | **0.001** |
| Dias descriptors | 2.019 | 2.552 | 0.439 | 0.001 |
| Blosum (Interface) | 2.055 | 2.626 | 0.433 | 0.002 |

Bold faced values indicate best performance for each model.
MSE: Mean Absolute Error, RMSE: Root Mean Sequred Error, $P_r$: Pearson correlation coefficient

of 2.090 kcal/mol (Table I). As shown in the scatter plot of predicted versus experimental binding affinities (Fig. 2(b)), the predictions exhibit a moderate linear association with the experimental values but with considerable dispersion around the identity line, particularly for high-affinity complexes. This spread is further reflected in the Bland–Altman analysis (Fig. 3(b)), where the prediction errors show substantial variability around the zero-bias line, with several points approaching or exceeding the predefined limits of agreement (±1.96(SD) kcal/mol), indicating limited agreement between predicted and experimental affinities.

The distribution of prediction errors (Fig. 4(b)) reveals a broad and slightly skewed error profile centered away from zero, suggesting both high variance and mild systematic bias in the sequence-based predictions. Consistently, the residual plot (Fig. 5(b)), depicting prediction error as a function of experimental binding affinity, shows heteroscedastic residuals and affinity-dependent trends, with larger errors observed for complexes with stronger binding. These complementary analyses indicate that although sequence-based models are broadly applicable, they struggle to fully capture the physicochemical determinants of protein–protein interactions, a limitation that has also been reported in previous studies.

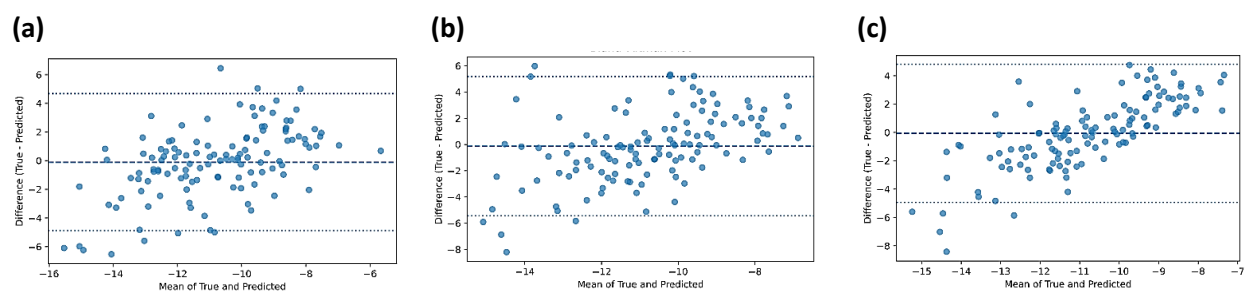

**Fig. 3. Bland–Altman plots illustrating agreement between predicted and experimental protein–protein binding affinities using propy and Moal descriptors across LOCO CV. (a)** structure-based baseline model, **(b)** sequence-based baseline model, and **(c)** knowledge distillation–based model. The bold dashed horizontal line represents the mean prediction error, while the thin dashed lines denote the 95% limits of agreement (±1.96 standard deviations). The sequence-based model shows larger bias and wider limits of agreement, whereas the structure-based model exhibits reduced error and improved agreement. The distillation-based student model demonstrates substantially reduced bias and narrower limits of agreement compared to the sequence-only baseline, indicating improved predictive accuracy and robustness through knowledge transfer.

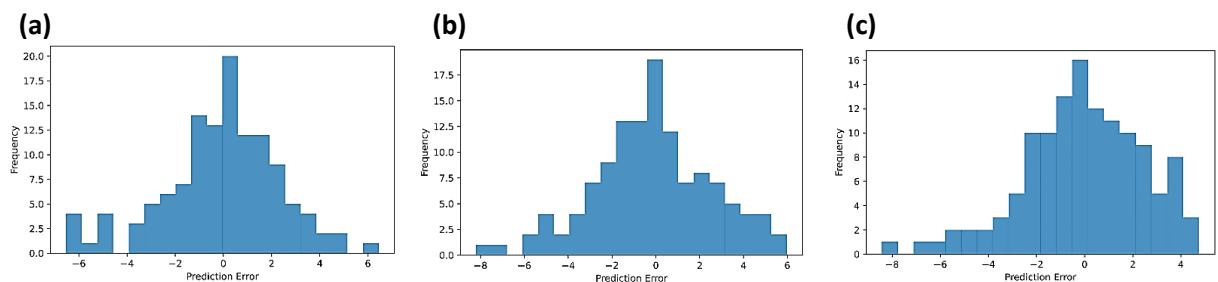

**Fig. 4.** Distribution of prediction errors (experimental minus predicted binding affinity) using propy and Moal descriptors across LOCO CV. **(a)** structure-based baseline model, **(b)** sequence-based baseline model, and **(c)** knowledge distillation–based model. The sequence-based model exhibits a broader and skewed error distribution, indicating higher variance and systematic inaccuracies. In contrast, the structure-based model shows a more concentrated error distribution around zero. The distillation-based student model demonstrates a narrower and more symmetric error distribution with reduced spread, reflecting improved predictive accuracy and robustness achieved through knowledge transfer.

**Baseline Binding Affinity Prediction Using Protein Structure Descriptors**

Next, we evaluated baseline regression models that were trained with structure-based descriptors that were taken from the bound and unbound three-dimensional conformations of protein complexes. Explicit spatial, energetic, and interfacial interaction information is encoded by these descriptors.

With a maximum Pearson correlation coefficient ($P_r$) of 0.512, an RMSE of 2.445 kcal/mol, and an MAE of 1.845 kcal/mol, structure-based models significantly outperformed sequence-only models using the LOCO cross-validation protocol (Table 2). The structure-based predictions are more closely clustered along the identity line than the sequence-based model, indicating improved linear agreement, as seen in the scatter plot of predicted versus experimental binding affinities (Fig. 3(a)). However, there is still discernible dispersion, especially at the extremes of the affinity range. The Bland-Altman analysis (Fig. 4(a)), which shows better but still imperfect agreement, further supports this improvement. Prediction errors are more symmetrically distributed around the zero-bias line, with fewer points approaching the predetermined limits of agreement (±1.96(SD) kcal/mol).

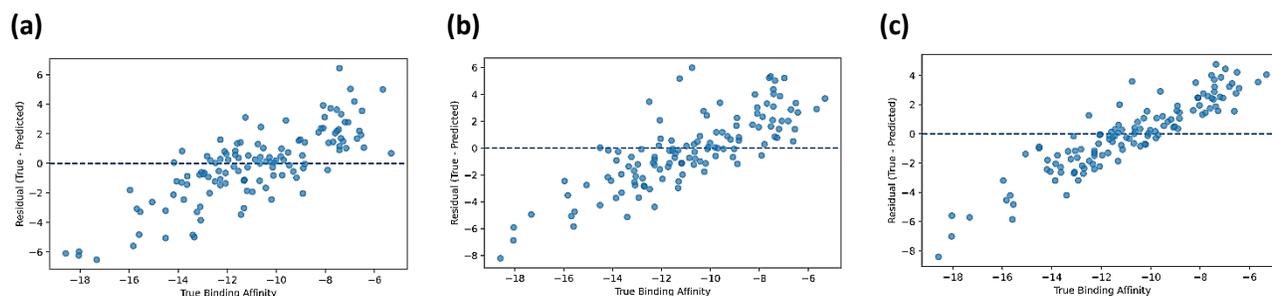

**Fig. 5.** Residual plots showing the relationship between prediction errors (experimental minus predicted binding affinity) and experimental binding affinities using propy and Moal descriptors across LOCO CV. **(a)** structure-based baseline model, **(b)** sequence-based baseline model, and **(c)** knowledge distillation–based model. The sequence-based model exhibits larger residual variance and noticeable trends across the affinity range, indicating systematic prediction errors. The structure-based model shows reduced residual spread with fewer systematic patterns. The distillation-based student model demonstrates more uniformly distributed residuals centered around zero, reflecting improved predictive accuracy and reduced bias through knowledge transfer.

The error distribution plot in Fig. 5(a) indicates that, in contrast to the sequence-only model, there exists a relatively narrower and more centralized error distribution, thereby reflecting a smaller variance and greater resistance to outliers. Nevertheless, slight skewness and noticeable spreading indicate that prediction error remains. This has been vindicated in that, by close inspection of Fig. 6(a), it can be found that, in contrast to the sequence-only model, better resistance to affine trends in addition to smaller heteroscedasticity exist in residuals, though greater variances in residuals emerge for biological complexes possessing highly variable and small binding affinities. These assessments capture that employment of structural features leads to improved predictive performance in binding affinity prediction in contrast to the sequence-only model, yet without achieving perfect performance owing to insufficiencies in training datasets.

**Knowledge Distillation–Based Protein Binding Affinity Prediction**

We assessed the suggested knowledge distillation framework, in which a structure-informed teacher network oversees the training of a sequence-based student network, in order to get around the drawbacks of sequence-only and structure-only approaches. In order to embed structural knowledge into a sequence-only predictor while preserving practical applicability at inference time, the student model simultaneously learns from the ground-truth binding affinity labels, the teacher's predictions, and intermediate feature representations during training.

The knowledge distillation-based models clearly outperformed sequence-only baselines using the LOCO cross-validation protocol. A maximum Pearson correlation coefficient ($P_r$) of 0.481, an RMSE of 2.488 kcal/mol, and an MAE of 1.927 kcal/mol were attained by the distillation-based student model (Table 3). This shows a significant improvement in predictive accuracy compared to the sequence-only baseline, with an improvement of roughly +0.106 in $P_r$ and a decrease of 0.224 kcal/mol in RMSE. The predicted binding affinities from the distillation-based model exhibit a notably stronger linear association with the experimental values than those from the sequence-only approach, as shown in Fig. 2(c), with predictions more closely spaced around the identity line. Improved agreement is further confirmed by the corresponding Bland-Altman plot (Fig. 3(c)), which shows narrower limits of agreement and a lower mean prediction bias when compared to sequence-only models. Additionally, the residual plot (Fig. 5(c)) displays less heteroscedasticity, and the prediction error distribution (Fig. 4(c)) is more sharply centered around zero, both of which point to improved model stability.

Although the suggested distillation framework greatly enhances sequence-based prediction performance, it falls short of structure-based models in terms of accuracy. The scarcity of training

**Table 3.** Performance of the distilled student regression model, where protein sequence descriptors are guided by a teacher trained on protein structure descriptors, evaluated using LOCO cross-validation over the affinity benchmark dataset.

| Descriptors (Teacher) | Descriptors (Srudent) | MSE | RMSE | $P_r$ | P-value |
|---|---|---|---|---|---|
| NIRP | k-mer | 2.149 | 2.676 | 0.335 | 0.002 |
| | k-mer-G | 1.988 | 2.582 | 0.412 | 0.001 |
| | Blosum | 2.118 | 2.626 | 0.397 | 0.001 |
| | Propy | 2.016 | 2.605 | 0.393 | 0.001 |
| | PSSM | 2.164 | 2.760 | 0.298 | 0.001 |
| | ProtParam | 2.182 | 2.708 | 0.327 | 0.002 |
| Moal descriptors | k-mer | 2.093 | 2.605 | 0.389 | 0.001 |
| | k-mer-G | 2.035 | 2.601 | 0.392 | 0.000 |
| | Blosum | 2.125 | 2.656 | 0.372 | 0.002 |
| | **Propy** | **1.927** | **2.488** | **0.481** | **0.001** |
| | PSSM | 2.208 | 2.765 | 0.300 | 0.001 |
| | ProtParam | 2.186 | 2.722 | 0.289 | 0.002 |
| Dias descriptors | k-mer | 2.081 | 2.616 | 0.311 | 0.001 |
| | k-mer-G | 2.013 | 2.554 | 0.372 | 0.002 |
| | Blosum | 2.145 | 2.680 | 0.394 | 0.001 |
| | Propy | 2.101 | 2.606 | 0.414 | 0.000 |
| | PSSM | 2.242 | 2.762 | 0.333 | 0.001 |
| | ProtParam | 2.209 | 2.738 | 0.295 | 0.002 |
| Blosum (Interface) | k-mer | 2.165 | 2.664 | 0.335 | 0.001 |
| | k-mer-G | 2.014 | 2.544 | 0.454 | 0.001 |
| | Blosum | 2.065 | 2.625 | 0.394 | 0.001 |
| | Propy | 2.043 | 2.573 | 0.421 | 0.001 |
| | PSSM | 2.182 | 2.748 | 0.330 | 0.001 |
| | ProtParam | 2.197 | 2.881 | 0.310 | 0.001 |

Bold faced values indicate best performance for each model
MSE: Mean Absolute Error, RMSE: Root Mean Sequred Error, $P_r$: Pearson correlation coefficient

data, which limits the efficient transfer of rich structural information from the teacher to the student network, is primarily responsible for this discrepancy. However, knowledge distillation is a workable and successful approach, as evidenced by the steady improvements seen across all evaluation metrics. Significantly, these findings imply that the suggested method may allow sequence-only models to attain performance levels comparable to structure-based predictors in the presence of sufficiently large and varied training datasets, thereby combining high accuracy with wide applicability.

**External Validation on Independent Protein–Protein Complexes**

We evaluated both the sequence-based baseline and the knowledge distillation-based student models on an independent external validation dataset that included 39 protein–protein complexes in order to further evaluate the generalization capability of the suggested framework. This dataset's complexes don't overlap with the benchmark dataset that was used to create the model. Using Moal descriptors for the teacher network and Propy descriptors for the student network, we only chose the top-performing models found under the LOCO cross-validation setting for this analysis. These models provide a rigorous and realistic evaluation of model robustness and practical applicability because they were trained on the entire non-redundant binding affinity benchmark dataset and then assessed on the external validation set.

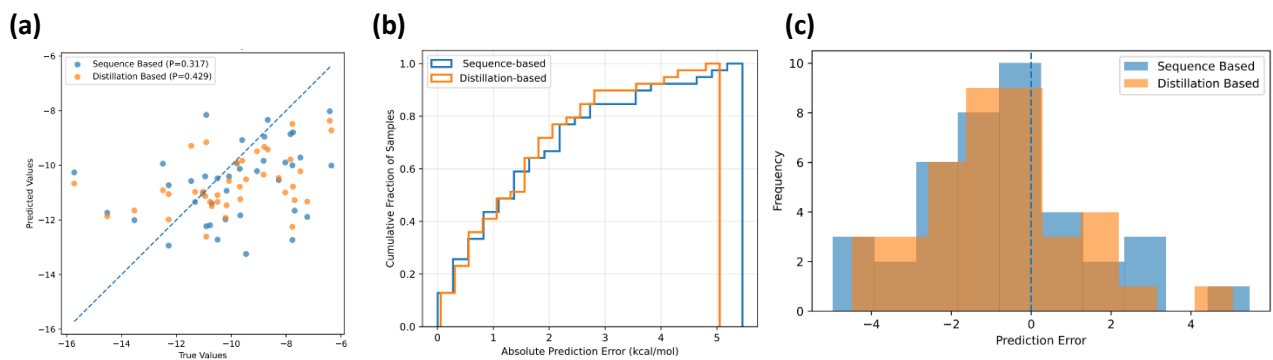

**Fig. 6. Performance comparison of sequence-based baseline and knowledge distillation–based student models on the external validation dataset of 39 non-overlapping protein–protein complexes. (a)** Performance comparison of sequence-based baseline and knowledge distillation–based student models on the external validation dataset of 39 non-overlapping protein–protein complexes. **(b)** Cumulative histogram of absolute prediction errors showing a higher proportion of low-error predictions for the distillation-based model compared to the sequence-only baseline. **(c)** Distribution of prediction errors highlighting reduced variance and improved error centralization for the distillation-based approach. All results are reported using the best-performing models selected under the LOCO cross-validation setting, with Moal structural descriptors used for the teacher network and Propy sequence descriptors used for the student network. Models were trained on the complete non-redundant binding affinity benchmark dataset and evaluated exclusively on the external validation set.

On the external validation dataset, the sequence-based baseline model achieved a Pearson correlation coefficient ($P_r$) of 0.317, an RMSE of 2.218 kcal/mol, and an MAE of 1.706 kcal/mol. As shown in the scatter plot of predicted versus experimental binding affinities (Fig. 6(a)), the predictions exhibit substantial dispersion around the identity line, particularly for high-affinity complexes, indicating limited generalization to unseen data. This behavior is further supported by the cumulative histogram of absolute prediction errors (Fig. 6(b)), where a relatively small fraction of samples achieves low-error predictions, reflecting weaker overall accuracy. Consistently, the distribution of prediction errors (Fig. 6(c)) is broad and mildly skewed suggests reduced robustness under distribution shift.

On the external dataset, however, the student model based on knowledge distillation shows better generalization. With a ($P_r$) of 0.429, an RMSE of 2.025 kcal/mol, and an MAE of 1.605 kcal/mol, the distillation-based model outperformed the sequence-only baseline by about +0.172 in correlation and 0.198 kcal/mol in RMSE(Table IV). Stronger agreement with experimental affinities is indicated by the distillation-based predictions being more closely clustered along the identity line, as shown in Fig. 6(a). This is again reinforced with the cumulative error histogram in Fig. 6(b), which shows that the plot for distillation is steeper, thereby ensuring that the number of complexes predicted with the lower absolute error is even higher. Also, the error distribution is centered around zero in Fig. 6(c), which again depicts better stability and robustness.

Notably, when tested on the same external validation dataset, the suggested distillation-based model also performed better than the previously published state-of-the-art sequence only approach trained on the same benchmark dataset (ISLAND), obtaining higher correlation and lower error metrics(Abbasi et al., 2020). Even in difficult generalization scenarios where structural information is not available at inference time, this comparison highlights the efficacy of incorporating structural knowledge into a sequence-only predictor.

Overall, these findings show that the suggested knowledge distillation framework not only increases cross-validation predictive accuracy but also more successfully generalizes to protein complexes that have never been seen before. The method's robustness is demonstrated by the steady improvements seen in scatter plots, cumulative error histograms, and error distribution analyses. The observed improvements show that the suggested approach is a significant step toward closing the performance gap, even though performance on the external dataset is still below ideal, probably because of insufficient training data and intrinsic experimental noise. Crucially, these results imply that the suggested framework has a good chance of achieving structure-level performance while maintaining the wide applicability of sequence-only predictors with bigger and more varied training datasets.

## Discussion

The intrinsic trade-off between predictive accuracy and data availability makes predicting protein–protein binding affinity a fundamental challenge in computational structural biology. By explicitly modeling interfacial geometry, energetic contributions, and spatial complementarity between interacting proteins, structure-based machine learning techniques have continuously shown superior performance(Gainza et al., 2020; Liu et al., 2021; Rodrigues et al., 2019; Romero-Molina et al., 2022; Wee and Xia, 2022). However, the lack of experimentally resolved protein structures severely restricts their practical applicability, especially for transient interactions and newly characterized proteins(Kastritis and Bonvin, 2013). Although sequence-based methods are widely applicable, their lack of direct knowledge of three-dimensional interactions and binding energetics frequently results in lower prediction accuracy (Abbasi and Minhas, 2016; Yugandhar and Gromiha, 2014). In this work, we propose a knowledge distillation-based regression framework that successfully converts structural knowledge into a sequence-only binding affinity predictor, thereby addressing this long-standing limitation.

Our findings demonstrate the modest predictive performance of baseline sequence-based models, which is in line with earlier research that found low correlation and comparatively high prediction error when depending only on sequence-derived descriptors. Significant dispersion, systematic bias, and heteroscedasticity are revealed by diagnostic analyses using scatter plots, Bland-Altman plots, prediction error distributions, and residual plots, especially for high-affinity complexes. These observations support previous findings that the physicochemical determinants of protein–protein interactions, which are intrinsically controlled by three-dimensional structural organization and interfacial contacts, are difficult for sequence-only representations to capture (Abbasi et al., 2020, 2018).

Structure-based models, on the other hand, demonstrated noticeably better predictive performance, supporting previous findings that structural descriptors offer more detailed and insightful depictions of binding energetics and interaction interfaces. The benefits of using explicit spatial and energetic features are demonstrated by the improved agreement, decreased error variance, and tighter clustering of predictions along the identity line seen in Bland-Altman analyses. However, structure-based models still do not achieve optimal prediction accuracy despite their superior performance. The intrinsic complexity of protein–protein interaction energetics, experimental noise in binding affinity measurements, and the comparatively small size of available benchmark datasets are probably the causes of this limitation (Charih et al., 2025). Further, their reliance on high-quality structural data significantly limits their applicability in large-scale or proteome-wide studies.

This could be achieved in a principled way with the proposed knowledge distillation framework, allowing a sequence-based student model to leverage structural information during training while being oblivious to it at inference time. Distillation-based models showed consistent improvements over sequence-only baselines both in Leave-One-Complex-Out cross-validation and external validation experiments due to improved correlation coefficients, smaller prediction errors, and enhanced stability. Notably, such gains emerged not only in aggregate performance metrics but also across a variety of diagnostic plots, including reduced bias in Bland–Altman analyses, narrower error distributions, and more homoscedastic residuals. These findings confirm the notion that the student network internalizes structurally informed representations from the teacher network via supervision and hence makes more robust predictions.

The results of the external validation experiments also highlight the application relevance of the proposed approach. When tested on a separate data set consisting of protein complexes that do not share anything with the training data set, the distillation approach still had a clear edge over the baseline method based solely on the sequence information. This confirms results observed in other similar studies pertaining to learning using privileged information(Abbasi et al., 2018).

Although such improvements have been achieved, the performance of the distillation-based student model has yet to be brought to the level of structure-based predictors. This can primarily be explained by the fact that the binding affinity datasets are, at present, too small in scale and scope. It has already been shown in previous works that deep learning models used for the prediction of molecular interactions clearly gain in performance with larger and more varied datasets (Liu et al., 2024). We thus expect that, with more and better binding affinity measurements being generated, the performance of our proposed distillation framework is supposed to further improve, and may even be capable of competing with structure-based methods based on model sequences alone.

The flexibility of the suggested framework is another benefit. The distillation-based approach separates training-time supervision from test-time requirements, in contrast to feature concatenation or hybrid models that require structural information at both training and inference time. Because of its design, it is especially appropriate for real-world situations where structural information is scarce but sequence data is abundant. Moreover, the framework is easily expandable to include more expressive structural encoders, such as graph-based and geometric deep learning architectures, or sophisticated sequence representations, like pretrained protein language models(Dandibhotla et al., 2025; Guo and Yamaguchi, 2022b).

In conclusion, this study shows that knowledge distillation provides a scalable and efficient method for incorporating protein sequence and structure information in binding affinity prediction. The suggested framework reduces the long-standing trade-off between predictive accuracy and applicability by incorporating structural knowledge into a sequence-based predictor. Knowledge distillation appears to be a promising approach for creating precise, reliable, and widely applicable protein–protein interaction predictors, despite ongoing difficulties with data scarcity and biological complexity.

## Conlusions and Future Work

We introduce a framework for protein-protein binding affinity prediction based on knowledge distillation that incorporates structural knowledge into a sequence-only predictor. The method uses three-dimensional descriptors during training and only needs sequence information during inference by training a student model under the direction of a structure-informed teacher network. The distillation-trained student consistently outperformed sequence-only baselines using Leave-One-Complex-Out cross-validation on a non-redundant benchmark dataset and an external validation dataset, demonstrating improved Pearson correlation, decreased RMSE, and increased prediction stability. More precise and reliable predictions than sequence-based models are confirmed by analyses of scatter plots, Bland-Altman plots, prediction error distributions, and residuals.

Because of the intricacy of protein–protein interaction energetics and the scarcity of structural data, student models still fall short of the accuracy of structure-based predictors despite these advancements. Expanding training datasets with experimentally resolved and predicted structures, incorporating cutting-edge neural architectures like attention-based and graph neural networks, and using multi-task learning to capture complementary biophysical properties are all areas of future research. All things considered, this distillation framework offers a workable and efficient method to enhance sequence-based binding affinity prediction, with great potential for extensive and proteome-wide applications.

## Availability of Data and Materials

All data generated or analyzed during this study are included in this paper or available at online repositories. A Python implementation of the proposed method is available at

https://sites.google.com/view/wajidarshad/software and https://github.com/wajidarshad/ProteinAffinityKD.

## Competing Interests

## Funding Statement

This study was not supported by any funding.

## Ethics Approval and Consent to Participate

This research does not involve human subjects, human material, or human data.

## Consent for Publication

This manuscript does not contain details, images, or videos relating to an individual person.

## References

Abbasi WA, Asif A, Ben-Hur A, et al. Learning Protein Binding Affinity Using Privileged Information. BMC Bioinformatics 2018;19(1):425; doi: 10.1186/s12859-018-2448-z.

Abbasi WA and Minhas FUAA. Issues in Performance Evaluation for Host–Pathogen Protein Interaction Prediction. J Bioinform Comput Biol 2016;14(03):1650011; doi: 10.1142/S0219720016500116.

Abbasi WA, Yaseen A, Hassan FU, et al. ISLAND: In-Silico Proteins Binding Affinity Prediction Using Sequence Information. BioData Min 2020;13(1):20; doi: 10.1186/s13040-020-00231-w.

Ahmad S and Mizuguchi K. Partner-Aware Prediction of Interacting Residues in Protein-Protein Complexes from Sequence Data. PLOS ONE 2011;6(12):e29104; doi: 10.1371/journal.pone.0029104.

Ain QU, Aleksandrova A, Roessler FD, et al. Machine-Learning Scoring Functions to Improve Structure-Based Binding Affinity Prediction and Virtual Screening. Wiley Interdiscip Rev Comput Mol Sci 2015;5(6):405–424; doi: 10.1002/wcms.1225.

Altschul SF, Madden TL, Schäffer AA, et al. Gapped BLAST and PSI-BLAST: A New Generation of Protein Database Search Programs. Nucleic Acids Res 1997;25(17):3389–3402.

Audie J and Scarlata S. A Novel Empirical Free Energy Function That Explains and Predicts Protein–Protein Binding Affinities. Biophys Chem 2007;129(2–3):198–211; doi: 10.1016/j.bpc.2007.05.021.

Aumentado-Armstrong TT, Istrate B and Murgita RA. Algorithmic Approaches to Protein-Protein Interaction Site Prediction. Algorithms Mol Biol AMB 2015;10:1–21; doi: 10.1186/s13015-015-0033-9.

Cao D-S, Xu Q-S and Liang Y-Z. Propy: A Tool to Generate Various Modes of Chou's PseAAC. Bioinformatics 2013;29(7):960–962; doi: 10.1093/bioinformatics/btt072.

Charih F, Green JR and Biggar KK. Sequence-Based Protein–Protein Interaction Prediction and Its Applications in Drug Discovery. Cells 2025;14(18):1449; doi: 10.3390/cells14181449.

Chen J, Sawyer N and Regan L. Protein–Protein Interactions: General Trends in the Relationship between Binding Affinity and Interfacial Buried Surface Area. Protein Sci Publ Protein Soc 2013;22(4):510–515; doi: 10.1002/pro.2230.

Dandibhotla S, Samudrala M, Kaneriya A, et al. GNNSeq: A Sequence-Based Graph Neural Network for Predicting Protein–Ligand Binding Affinity. Pharmaceuticals 2025;18(3):329; doi: 10.3390/ph18030329.

Dias R and Kolaczkowski B. Improving the Accuracy of High-Throughput Protein-Protein Affinity Prediction May Require Better Training Data. BMC Bioinformatics 2017;18(Suppl 5); doi: 10.1186/s12859-017-1533-z.

Eddy SR. Where Did the BLOSUM62 Alignment Score Matrix Come From? Nat Biotechnol 2004;22(8):1035–1036; doi: 10.1038/nbt0804-1035.

Gainza P, Sverrisson F, Monti F, et al. Deciphering Interaction Fingerprints from Protein Molecular Surfaces Using Geometric Deep Learning. Nat Methods 2020;17(2):184–192; doi: 10.1038/s41592-019-0666-6.

Gasteiger E, Hoogland C, Gattiker A, et al. Protein Identification and Analysis Tools on the ExPASy Server. In: The Proteomics Protocols Handbook. (Walker JohnM. ed) Humana Press; 2005; pp. 571–607; doi: 10.1385/1-59259-890-0:571.

Guo Z and Yamaguchi R. Machine Learning Methods for Protein-Protein Binding Affinity Prediction in Protein Design. Front Bioinforma 2022a;2.

Guo Z and Yamaguchi R. Machine Learning Methods for Protein-Protein Binding Affinity Prediction in Protein Design. Front Bioinforma 2022b;2; doi: 10.3389/fbinf.2022.1065703.

Guruprasad K, Reddy BV and Pandit MW. Correlation between Stability of a Protein and Its Dipeptide Composition: A Novel Approach for Predicting in Vivo Stability of a Protein from Its Primary Sequence. Protein Eng 1990;4(2):155–161.

Harris D and Harris S. Digital Design and Computer Architecture. Second Edition. Morgan Kaufmann: Amsterdam; 2012.

Hinton G, Vinyals O and Dean J. Distilling the Knowledge in a Neural Network. 2015; doi: 10.48550/arXiv.1503.02531.

Horton N and Lewis M. Calculation of the Free Energy of Association for Protein Complexes. Protein Sci Publ Protein Soc 1992;1(1):169–181.

Huggins DJ, Sherman W and Tidor B. Rational Approaches to Improving Selectivity in Drug Design. J Med Chem 2012;55(4):1424–1444; doi: 10.1021/jm2010332.

Hwang H, Vreven T, Janin J, et al. Protein-Protein Docking Benchmark Version 4.0. Proteins 2010;78(15):3111–3114; doi: 10.1002/prot.22830.

Kairys V, Baranauskiene L, Kazlauskiene M, et al. Binding Affinity in Drug Design: Experimental and Computational Techniques. Expert Opin Drug Discov 2019;14(8):755–768; doi: 10.1080/17460441.2019.1623202.

Kastritis PL and Bonvin AMJJ. On the Binding Affinity of Macromolecular Interactions: Daring to Ask Why Proteins Interact. J R Soc Interface 2013;10(79):20120835; doi: 10.1098/rsif.2012.0835.

Kastritis PL, Moal IH, Hwang H, et al. A Structure-Based Benchmark for Protein-Protein Binding Affinity. Protein Sci Publ Protein Soc 2011;20(3):482–491; doi: 10.1002/pro.580.

Leslie C, Eskin E and Noble WS. The Spectrum Kernel: A String Kernel for SVM Protein Classification. Pac Symp Biocomput Pac Symp Biocomput 2002;7:564–575.

Li ZR, Lin HH, Han LY, et al. PROFEAT: A Web Server for Computing Structural and Physicochemical Features of Proteins and Peptides from Amino Acid Sequence. Nucleic Acids Res 2006;34(suppl 2):W32–W37; doi: 10.1093/nar/gkl305.

Limongelli I, Marini S and Bellazzi R. PaPI: Pseudo Amino Acid Composition to Score Human Protein-Coding Variants. BMC Bioinformatics 2015;16:123; doi: 10.1186/s12859-015-0554-8.

Liu H, Chen P, Zhai X, et al. PPB-Affinity: Protein-Protein Binding Affinity Dataset for AI-Based Protein Drug Discovery. Sci Data 2024;11(1):1316; doi: 10.1038/s41597-024-03997-4.

Liu X, Luo Y, Li P, et al. Deep Geometric Representations for Modeling Effects of Mutations on Protein-Protein Binding Affinity. PLOS Comput Biol 2021;17(8):e1009284; doi: 10.1371/journal.pcbi.1009284.

Marsh JA and Teichmann SA. Structure, Dynamics, Assembly, and Evolution of Protein Complexes. Annu Rev Biochem 2015;84:551–575; doi: 10.1146/annurev-biochem-060614-034142.

Moal IH, Agius R and Bates PA. Protein-Protein Binding Affinity Prediction on a Diverse Set of Structures. Bioinformatics 2011;btr513; doi: 10.1093/bioinformatics/btr513.

Moal IH and Fernández-Recio J. SKEMPI: A Structural Kinetic and Energetic Database of Mutant Protein Interactions and Its Use in Empirical Models. Bioinforma Oxf Engl 2012;28(20):2600–2607; doi: 10.1093/bioinformatics/bts489.

Moal IH and Fernández-Recio J. Comment on 'Protein–Protein Binding Affinity Prediction from Amino Acid Sequence.' Bioinformatics 2014;btu682; doi: 10.1093/bioinformatics/btu682.

Moerke NJ. Fluorescence Polarization (FP) Assays for Monitoring Peptide-Protein or Nucleic Acid-Protein Binding. Curr Protoc Chem Biol 2009;1(1):1–15; doi: https://doi.org/10.1002/9780470559277.ch090102.

Panday SK and Alexov E. Protein-Protein Binding Free Energy Predictions with the MM/PBSA Approach Complemented with the Gaussian-Based Method for Entropy Estimation. ACS Omega 2022;7(13):11057–11067; doi: 10.1021/acsomega.1c07037.

Paszke A, Gross S, Massa F, et al. PyTorch: An Imperative Style, High-Performance Deep Learning Library. 2019; doi: 10.48550/arXiv.1912.01703.

Pruitt KD, Tatusova T and Maglott DR. NCBI Reference Sequence (RefSeq): A Curated Non-Redundant Sequence Database of Genomes, Transcripts and Proteins. Nucleic Acids Res 2005;33(suppl 1):D501–D504; doi: 10.1093/nar/gki025.

Rodrigues CHM, Myung Y, Pires DEV, et al. mCSM-PPI2: Predicting the Effects of Mutations on Protein–Protein Interactions. Nucleic Acids Res 2019;47(W1):W338–W344; doi: 10.1093/nar/gkz383.

Romero-Molina S, Ruiz-Blanco YB, Mieres-Perez J, et al. PPI-Affinity: A Web Tool for the Prediction and Optimization of Protein–Peptide and Protein–Protein Binding Affinity. J Proteome Res 2022;21(8):1829–1841; doi: 10.1021/acs.jproteome.2c00020.

Saponaro A. Isothermal Titration Calorimetry: A Biophysical Method to Characterize the Interaction between Label-Free Biomolecules in Solution. Bio-Protoc 2018;8(15):e2957; doi: 10.21769/BioProtoc.2957.

Sekar RB and Periasamy A. Fluorescence Resonance Energy Transfer (FRET) Microscopy Imaging of Live Cell Protein Localizations. J Cell Biol 2003;160(5):629–633; doi: 10.1083/jcb.200210140.

Shen J, Zhang J, Luo X, et al. Predicting Protein–Protein Interactions Based Only on Sequences Information. Proc Natl Acad Sci U S A 2007;104(11):4337–4341; doi: 10.1073/pnas.0607879104.

Siebenmorgen T and Zacharias M. Computational Prediction of Protein–Protein Binding Affinities. Wiley Interdiscip Rev Comput Mol Sci 2019; doi: 10.1002/wcms.1448.

Su Y, Zhou A, Xia X, et al. Quantitative Prediction of Protein-Protein Binding Affinity with a Potential of Mean Force Considering Volume Correction. Protein Sci Publ Protein Soc 2009;18(12):2550–2558; doi: 10.1002/pro.257.

Swapna LS, Bhaskara RM, Sharma J, et al. Roles of Residues in the Interface of Transient Protein-Protein Complexes before Complexation. Sci Rep 2012;2:334; doi: 10.1038/srep00334.

Tang Y, Zeng X and Liang J. Surface Plasmon Resonance: An Introduction to a Surface Spectroscopy Technique. J Chem Educ 2010;87(7):742–746; doi: 10.1021/ed100186y.

Wee J and Xia K. Persistent Spectral Based Ensemble Learning (PerSpect-EL) for Protein–Protein Binding Affinity Prediction. Brief Bioinform 2022;23(2):bbac024; doi: 10.1093/bib/bbac024.

Westen GJ, Swier RF, Wegner JK, et al. Benchmarking of Protein Descriptor Sets in Proteochemometric Modeling (Part 1): Comparative Study of 13 Amino Acid Descriptor Sets. J Cheminformatics 2013;5(1):41; doi: 10.1186/1758-2946-5-41.

Yugandhar K and Gromiha MM. Protein-Protein Binding Affinity Prediction from Amino Acid Sequence. Bioinformatics 2014;30(24):3583–3589; doi: 10.1093/bioinformatics/btu580.

Zaki N, Lazarova-Molnar S, El-Hajj W, et al. Protein-Protein Interaction Based on Pairwise Similarity. BMC Bioinformatics 2009;10(1):150; doi: 10.1186/1471-2105-10-150.

Zhu H, Sommer I, Lengauer T, et al. Alignment of Non-Covalent Interactions at Protein-Protein Interfaces. PLoS ONE 2008;3(4):e1926; doi: 10.1371/journal.pone.0001926.

# Supplementary Data

# Investigating Knowledge Distillation Through Neural Networks for Protein Binding Affinity Prediction

Wajid Arshad Abbasi[1, *], Syed Ali Abbas[1], Maryum Bibi[1], Saiqa Andleeb[2], Muhammad Naveed Akhtar[3]

[1]Computaional Biology and Data Analysis Lab., Institute of Computing, King Abdullah Campus, University of Azad Jammu & Kashmir, Muzaffarabad, AJ&K, 13100, Pakistan.

[2]Biotechnology Lab., Department of Zoology, King Abdullah Campus, University of Azad Jammu & Kashmir, Muzaffarabad, AJ&K, 13100, Pakistan.

[3]Department of Computer and Information Sciences(DCIS), Pakistan Institute of Engineering and Applied Sciences(PIEAS), Islamabad, Pakistan.

**\*Corresponding author**: Wajid A. Abbasi (e-mail: wajidarshad@uajk.edu.pk).

# Additional file 1

## Contents

**Table S1. Summary of Teacher and Student Network Architectures.** Architectural comparison of teacher and student neural networks used in the proposed knowledge distillation framework. The teacher model leverages structure-based descriptors, while the student model relies solely on sequence-based descriptors and is used during inference.

| Component | Teacher Network (Structure-Based) | Student Network (Sequence-Based) |
| --- | --- | --- |
| *Input descriptors* | Protein 3D structure–based features | Protein sequence–based features |
| *Input dimension* | ($d_t$) (structure descriptor–dependent) | ($d_s$) (sequence descriptor–dependent) |
| *Model type* | Fully connected feedforward neural network | Fully connected feedforward neural network |
| *Hidden layers* | Multiple linear layers with nonlinear activations | Multiple linear layers with nonlinear activations |
| *Activation functions* | ReLU (or equivalent) | ReLU (or equivalent) |
| *Distillation layer* | Selected intermediate hidden layer | Corresponding intermediate hidden layer |
| *Latent feature dimension* | Fixed dimension ($d_h$) | Fixed dimension ($d_h$) |
| *Output layer* | Linear regression head | Linear regression head |
| *Output* | Predicted binding affinity | Predicted binding affinity |
| *Training objective* | Supervised regression loss | Supervised(baseline) and Supervised + output-level + feature-level distillation losses(distillation) |
| *Training mode* | Jointly optimized (not frozen) | Jointly optimized |
| *Inference requirement* | Structure descriptors | Sequence descriptors only |

**Table S2. Neural Network Architectures.** Detailed architectural configurations for the teacher and student neural networks used in the proposed knowledge distillation framework.

| Component | Teacher Network (Structure-Based) | Student Network (Sequence-Based) |
|---|---|---|
| **Input dimension** | ($d_t$) (e.g., 20-200, descriptor-dependent) | ($d_s$) (e.g., 400–8000, descriptor-dependent) |
| **Hidden Layer 1** | Linear($d_t$, int(min(512,max(64,input_dim/8)))) + ReLU | Linear($d_t$, int(min(512,max(64,input_dim/8)))) + ReLU |
| **Hidden Layer 2** | Linear($d_{h1}$, int(min(128, h1/2))) + ReLU | Linear($d_{h1}$, int(min(128, h1/2))) + ReLU |
| **Distillation layer** | Linear($d_{h2}$, 16) + ReLU | Linear($d_{h2}$, 16) + ReLU |
| **Output layer** | Linear(16, 1) | Linear(16, 1) |
| **Output activation** | None (linear regression) | None (linear regression) |

Note: The distillation layer dimensionality was fixed to 64 across all models to enable feature-level alignment. Layer sizes were selected empirically to balance model capacity and overfitting risk.

**Table S3. Training Hyperparameters.** Detailed training hyperparameters for the teacher and student neural networks used in the proposed knowledge distillation framework.

| Hyperparameter | Value |
| --- | --- |
| **Framework** | PyTorch |
| **Optimizer** | Adam |
| **Learning rate** | $10^{-3}$ |
| **Weight decay (L2)** | $10^{-4}$ |
| **Batch size** | 1 |
| **Number of epochs** | 100 |
| **Loss function (supervised)** | Mean Squared Error (MSE) |
| **Output-level distillation weight ($\lambda_{out}$)** | 0.6 |
| **Feature-level distillation weight ($\lambda_{feat}$)** | 0.5 |
| **Teacher freezing** | No (joint optimization) |
| **Feature normalization** | StandardScaler (z-score normalization) |
| **Initialization** | Kaiming uniform (by default) |
| **Dropout** | 0.3(h1) and 0.2(h2) |